\documentclass[11pt,a4paper,final]{article}
\usepackage[left=1in,right=1in,top=1in,bottom=1in]{geometry}
\usepackage{authblk}
\PassOptionsToPackage{sort&compress}{natbib}
\usepackage[numbers]{natbib}
\usepackage{booktabs} 
\usepackage[ruled]{algorithm2e} 
\usepackage{tikz}
\usetikzlibrary{arrows}
\usetikzlibrary{shapes.multipart}
\usepackage{caption}

\SetAlFnt{\small}
\SetAlCapFnt{\small}
\SetAlCapNameFnt{\small}
\SetAlCapHSkip{0pt}
\IncMargin{-\parindent}
\usepackage[all]{nowidow}

\usepackage[font=small,labelfont=bf]{caption}
\usepackage[utf8]{inputenc} 
\usepackage[T1]{fontenc}    
\usepackage{hyperref}       
\hypersetup{
    colorlinks=false,
    linkcolor=blue,
    filecolor=magenta,      
    urlcolor=cyan,
    pdfpagemode=FullScreen,
    }
\usepackage{url}            
\usepackage{booktabs}       
\usepackage{amsfonts}       
\usepackage{nicefrac}       
\usepackage{microtype}      
\usepackage{xcolor}         
\usepackage{amsmath}
\usepackage{mathtools}
\usepackage{amsthm}
\usepackage{amsmath}
\usepackage{amssymb}
\usepackage{bbm}
\usepackage{tikz} 
\usepackage{wrapfig}
\usepackage{floatrow}
\usepackage{enumitem}

\usepackage[suppress]{color-edits}
\addauthor[Hoda]{hh}{cyan}
\addauthor{kh}{red}
\addauthor{vc}{blue}
\addauthor{jk}{orange}
\addauthor{sw}{blue}

\usepackage{caption}
\usepackage{subcaption}

\DeclareMathOperator*{\argmin}{arg\,min}

\title{Perspectives on Incorporating Expert Feedback into \\ Model Updates}

\newcommand*\samethanks[1][\value{footnote}]{\footnotemark[#1]}

\begin{document}

\author[1]{Valerie Chen\thanks{Both authors contributed equally. Order was decided by a coin flip. Correspondence to: \href{mailto:valeriechen@cmu.edu}{valeriechen@cmu.edu}  and \href{mailto:usb20@cam.ac.uk}{usb20@cam.ac.uk} }}
\author[2,3]{Umang Bhatt\samethanks}
\author[1]{Hoda Heidari}
\author[2,3]{Adrian Weller}
\author[1]{Ameet Talwalkar}
\affil[1]{Carnegie Mellon University}
\affil[2]{University of Cambridge}
\affil[3]{The Alan Turing Institute}

\date{}

\maketitle

\begin{abstract}
Machine learning (ML) practitioners are increasingly tasked with developing models that are aligned with non-technical experts' values and goals. However, there has been insufficient consideration of how practitioners should translate domain expertise into ML updates. In this paper, we consider how to capture interactions between practitioners and experts systematically. 
We devise a taxonomy to match expert feedback types with practitioner updates.
A practitioner may receive feedback from an expert at the observation- or domain-level, and convert this feedback into updates to the dataset, loss function, or parameter space. 
We review existing work from ML and human-computer interaction to describe this feedback-update taxonomy, and highlight the insufficient consideration given to incorporating feedback from non-technical experts.
We end with a set of open questions that naturally arise from our proposed taxonomy and subsequent survey.
\end{abstract}

\section{Introduction}\label{sec:intro}
\begin{figure}[h!]
\centering
  \includegraphics[scale=0.95]{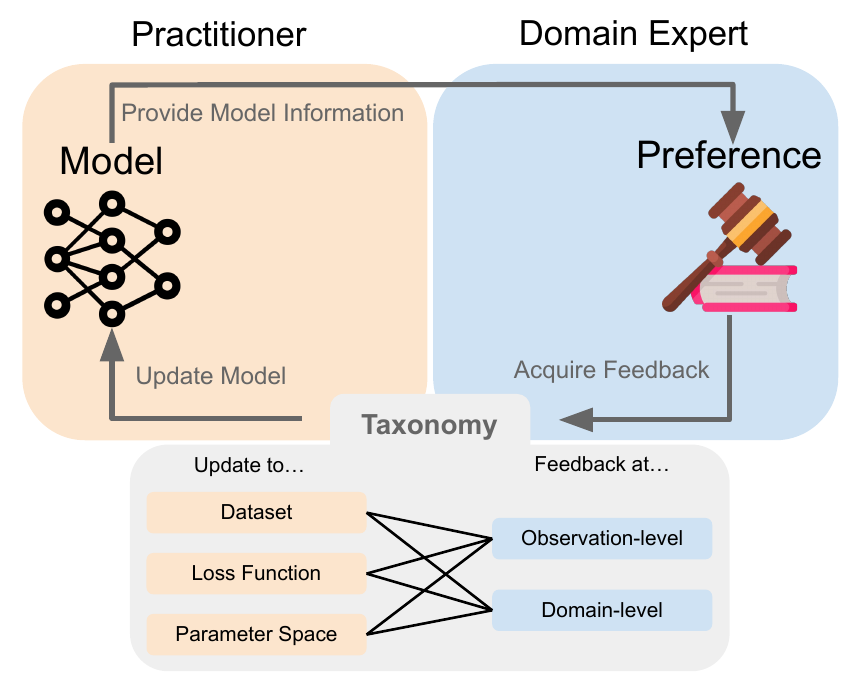}
  \caption{To incorporate an expert's preferences to improve a model, practitioners must turn non-technical, domain expert preferences into usable model updates. In this work, we propose a feedback-update taxonomy that focuses on the ways in which expert feedback can be translated into model updates. Our taxonomy has two axes, expert feedback and model updates, forming six categories of feedback-update interaction. We map existing work onto each category, and use our taxonomy to motivate improving the interaction between practitioners and domain experts.
  }
  \label{fig:example}
\end{figure}

Before deploying a machine learning (ML) model in high-stakes use cases, \emph{practitioners}, who are responsible for developing and maintaining models, may solicit and incorporate feedback from \emph{experts}~\cite{fails2003interactive,amershi2014power,cui2021understanding}.
Prior work has largely focused on incorporating feedback of technical experts (herein ML engineers, data scientists, etc.) into models~\cite{adebayo2020debugging,li2021intermittent,liu2017iterative,ross2017right,simard2017machine,song2013stochastic,wang2020mathematical}. The feedback of technical experts might be immediately actionable, as likely few communication barriers exist between technical experts and practitioners.
In contrast, the relationship between a practitioner and non-technical expert (herein doctors, lawyers, elected officials, policymakers, social workers, etc.), as illustrated in Figure~\ref{fig:example}, is more complex~\cite{bhatt2020explainable,chen2022interpretable}. 
Upon seeing information about the model, the expert provides feedback based on their preference to practitioners, who can then update the model. There has been insufficient consideration on how to incorporate feedback from non-technical, domain experts~\cite{bhatt2020machine,paml} into  models.







\begin{table*}[]
\centering
\small
\caption{Our feedback-update taxonomy illustrates the diverse ways practitioners can convert expert feedback, which generally comes via domain- or observation-level feedback, into model updates, which are either dataset, loss function, or parameter space changes that entail changes to $\mathcal{D}$, $\mathcal{L}$, or $\Theta$ respectively. 
Each cell corresponds to a subsection of Section~\ref{sec:taxonomy}. 
}
\resizebox{\textwidth}{!}{\begin{tabular}{llll}
& \multicolumn{1}{c}{Dataset Update} & \multicolumn{1}{c}{Loss Function Update} &\multicolumn{1}{c}{Parameter Space Update}   \\ \cline{2-4} 
\multicolumn{1}{l|}{\begin{tabular}[c]{@{}l@{}} \rotatebox{90}{Domain Feedback}
\end{tabular}}  & \multicolumn{1}{l|}
{\begin{tabular}[c]{@{}l@{}}
\textit{Dataset modification}\\
\qquad Augmentation~\citep{dao2019kernel}\\ 
\qquad Preprocessing~\citep{hajian2012methodology,feldman2015certifying, calmon2017optimized, iyengar2002transforming} \\ 
\textit{Data generation from constraint}\\
\qquad Fairness~\citep{xu2018fairgan}\\
\qquad Interpretability~\citep{plumb2021finding, lee2020explanation}\\
\textit{Weak supervision}\\
\qquad Using unlabeled data~\citep{ratner2017snorkel,carmon2019unlabeled,arazo2020pseudo}\\
\qquad Checking synthetic data~\citep{ryazanov2021deep}\\
\end{tabular}} & \multicolumn{1}{l|}
{\begin{tabular}[c]{@{}l@{}} 
\textit{Constraint specification}\\
\qquad Fairness~\citep{zafar2017fairness,dimanov2020you,hiranandani2020fair,coston2021characterizing}\\
\qquad Interpretability~\citep{lakkaraju2016interpretable, plumb2020regularizing}\\
\qquad Resource constraints~\citep{frankle2018lottery,lin2020mcunet}
\end{tabular}} 
& \multicolumn{1}{l|}
{\begin{tabular}[c]{@{}l@{}}
\textit{Model editing}\\
\qquad Rules~\citep{yang2019study}\\
\qquad Weights~\citep{wang2021gam}\\
\textit{Model selection}\\
\qquad Prior update~\citep{lage2018human}\\
\qquad Complexity~\citep{rudin2019stop,dziugaite2020enforcing}
\end{tabular}} 
\\ \cline{2-4} 
\multicolumn{1}{l|}{\begin{tabular}[c]{@{}l@{}}\rotatebox{90}{Observation Feedback}\end{tabular}} & \multicolumn{1}{l|}
{\begin{tabular}[c]{@{}l@{}} 
\textit{Active data collection}\\
\qquad Adding data~\citep{wan2020human,fanton2021human, cabrera2021discovering,ghai2020explainable}\\
\qquad Relabeling data~\citep{kaushik2019learning}\\
\qquad Reweighting data~\citep{bourtoule2021machine,zhao2019metric,hiranandani2021optimizing}\\
\qquad Collecting expert labels~\citep{peterson2019human}\\
\textit{Passive observation}~\citep{irvin2019chexpert, laidlaw2021uncertain, swartz2006inverse}\\
\end{tabular}} 
& \multicolumn{1}{l|}
{\begin{tabular}[c]{@{}l@{}}
\textit{Constraint elicitation}\\
\qquad Metric learning~\citep{hiranandani2019performance, jung2019algorithmic,cheng2021soliciting, yaghini2021human}\\
\qquad Human representations~\cite{hilgard2021learning, santurkar2021editing} \\
\textit{Collecting contextual information}\\
\qquad Generative factors~\cite{adel2018discovering}\\
\qquad Concept representations~\citep{koh2020concept,lage2020learning}\\
\qquad Explanations~\citep{hind2019ted}\\
\qquad Feature attributions~\citep{weinberger2020learning,tseng2020fourier}
\end{tabular}} 
& \multicolumn{1}{l|}
{\begin{tabular}[c]{@{}l@{}}
\textit{Feature modification}\\
\qquad Add/remove features~\citep{correia2019human,noriega2019active,bakker2021beyond}\\
\qquad Engineering features~\citep{lou2013accurate}
\end{tabular}} 
\\ \cline{2-4} 
\end{tabular}}
\label{tab:conversion}
\end{table*}

To bridge this gap, we start by examining model updates available to the practitioner and the types of feedback that non-technical experts might provide. We clarify the mechanisms available to turn feedback into updates and then devise a taxonomy along two axes: (a) levels of expert feedback, and (b) types of model updates. Along the first axis (Section~\ref{sec:feedback}), expert feedback may come as \emph{domain-level feedback}, which captures high-level conceptual feedback that the practitioner must translate into updates, or \emph{observation-level feedback}, which captures how the model should behave on a few, specific datapoints~\citep{hertwig2009description,davies2005inverse,swartz2006inverse,armstrong2018occam}. 
Along the other axis (Section~\ref{sec:model_updates}), we consider the updates a practitioner can make to a supervised learning objective, where feedback typically changes the dataset, the loss function, or the parameter space.\footnote{Supervised learning covers a broad range of model classes, ranging from vision transformers, large language models, and impactful application areas, like medical diagnostics~\cite{irvin2019chexpert} and criminal justice~\cite{pierson2020large}.
We consider other objectives, which may include reinforcement learning~\citep{christiano2017deep,cui2021understanding} or unsupervised learning~\citep{coden2017method,guimaraes2020human}, to be out-of-scope for this paper. We focus on methods more commonly deployed in practice, and omit the Bayesian analog for our pipeline, where experts can express preferences over the distribution of functions~\cite{guo2010gaussian, o2019expert}.} The two axes of our taxonomy form six distinct categories for feedback-update interactions. We place existing techniques from human-computer interaction, ML, robotics, and FATE (Fairness, Accountability, Transparency, Ethics) into each of these categories in Section~\ref{sec:taxonomy}.    

Our taxonomy not only provides a preliminary understanding on the ways in which non-technical expert feedback can be converted into practical model updates, but also motivates a diverse set of open questions to improve practitioner-expert interactions. 
In Section~\ref{sec:open}, we pose questions on how to connect model information to our taxonomy, how to prompt and elicit expert feedback effectively, and how to decide the type of update to perform given feedback.
We hope our taxonomy grounds the community in concrete ways to leverage non-technical expert feedback in a practical way while still encouraging future research to further feedback incorporation.




\section{Feedback-Update Taxonomy}\label{sec:pipeline}
One role of practitioners is to convert non-technical expert feedback into a model update.\footnote{We note the important case that sometimes valuable expert feedback might be received to say that using \emph{any} model is not appropriate for the setting at hand. While such a concern must be taken seriously and considered carefully with relevant stakeholders, we do not discuss this case further here. While experts are often involved prior to training an initial model~\cite{irvin2019chexpert}, we focus on the iterative feedback process after a model has been trained.} 
We describe the diverse ways that expert feedback 
can lead to model updates through our \emph{feedback-update taxonomy} (Table~\ref{tab:conversion}).%
\footnote{One piece of feedback could be used to alter multiple parts of the objective (e.g., change the dataset and loss function). Each update should be considered individually.} We first elaborate on the two axes of our taxonomy and flesh out each category in the next section.

\subsection{Levels of Domain Expert Feedback} \label{sec:feedback}
Once an expert has observed information about the model, practitioners may ask for feedback to improve the model's behavior in two general ways~\cite{hertwig2009description,wulff2018meta}. 

\begin{itemize}
    \item \textbf{Domain feedback.} It may be natural for non-technical experts to provide high-level conceptual feedback~\citep{hertwig2009description}. The expert could provide explicit feedback over a set of good models~\cite{fisher2019all,semenova2019study} or suggest data pre-processing to reduce discrimination \citep{hajian2012methodology,feldman2015certifying, calmon2017optimized}.
    \item \textbf{Observation feedback.} It may also be possible to learn by observing expert behavior~\citep{davies2005inverse,swartz2006inverse,armstrong2018occam}. For example, practitioners can use observations to approximate a property of interest (e.g., fairness)~\cite{hiranandani2020fair}, or can collect \emph{contextual information}, where every datapoint $(x_i, y_i)$ is accompanied by auxiliary information $s_i$ that can be used during learning (e.g., feature attributions~\cite{weinberger2020learning}, style factors~\cite{adel2018discovering}, semantically meaningful concepts~\cite{koh2020concept}).
\end{itemize}


These two types of feedback form one axis of our feedback-update taxonomy in Table~\ref{tab:conversion}. While these two forms of feedback may be non-exhaustive, they capture a wide variety of mechanisms for non-technical experts to influence the development of models~\cite{hertwig2009description}. Neither type requires non-technical experts to have knowledge about the model or training process itself. For example, a radiologist could provide ground-truth annotations on X-ray scans, which is observation-level feedback, or provide high-level information about the region of interest, which is domain-level feedback. 
The role of practitioners may be expanding. One unnatural thing a practitioner may need to do in addition to making the technical changes to the model is deciding whether the collected feedback is domain- or observation-level feedback given arbitrary expert feedback. 

We consider other forms of feedback to be out-of-scope for this work because they are less intuitive to elicit from a non-technical domain expert~\citep{schoeffer2021ranking,wang2020deontological}. This includes changing the learning algorithm (e.g., in differential privacy communities~\citep{song2013stochastic,dwork2014algorithmic}), selecting hyperparameters (e.g., in AutoML research~\citep{li2017hyperband, li2021intermittent}), and specifying the order of datapoints given to a learning algorithm (e.g., in machine teaching literature~\cite{simard2017machine,liu2017iterative,wang2020mathematical}).


\subsection{Types of Model Updates}\label{sec:model_updates}
In the supervised learning setting, a practitioner minimizes a loss function $\mathcal{L}(x,y; \theta)$ on a dataset $\mathcal{D} = \{(x_i, y_i)\}_{i=1}^n$ to learn a model parametrized by $\theta \in \Theta$:
\begin{equation*}\label{og}
    \hat{\theta} = \argmin_{\theta \in \Theta} \sum_{(x,y) \in \mathcal{D}} \mathcal{L}(x,y;\theta).
\end{equation*}
Once experts have provided feedback, practitioners can leverage expert input to improve the model in multiple ways: updating the dataset, the loss function, or the parameter space. These update types form the other axis of Table~\ref{tab:conversion}. 

\begin{itemize}
    \item \noindent\textbf{Dataset updates.} Feedback can be incorporated by changing the dataset from $\mathcal{D}$ to $\mathcal{D}^{\prime}$.
    \item \noindent\textbf{Loss function updates.} Feedback can also be incorporated by adding a constraint to the optimization objective. This manifests as a change in the loss function from $\mathcal{L}$ to $\mathcal{L}^{\prime}$.
    \item \noindent\textbf{Parameter space updates.} Finally, feedback can also be provided on the parameters of the model itself. These changes manifest as changing the parameter space in some sense, from $\Theta$ to  $\Theta^\prime$. 
\end{itemize}




For example, a public official may ask that as the input feature \texttt{population} increases, the likelihood of a project proposal getting \texttt{funded} should increase; this implies monotonicity between an outcome and input feature.
Practitioners can incorporate this feedback in various ways which are different technically. 
The practitioner can \textit{update the dataset} by adding or removing appropriate datapoints, \textit{update the loss function} by adding a regularizer that penalizes the model for not satisfying this condition, or \textit{update the parameter space} by optimizing over a subspace of parameters that satisfy this condition.

While these update types may seem straightforward, it is unclear how to find  $\mathcal{D}^\prime$, $\mathcal{L}^\prime$, or $\Theta^\prime$: this involves transforming domain expert feedback into one of these three general updates. After the practitioner incorporates expert feedback, the new updated model, $\tilde{\theta}$, should ideally reflect the expert's preferences better than the original model, $\hat{\theta}$. 
In the next section, we flesh out the \emph{conversion} from feedback to update.

\section{Mapping Prior Work to our Taxonomy}\label{sec:taxonomy}
For each feedback-update pair, we illustrate how the practitioner can use the feedback to modify a model parameterized by $\hat\theta$ into a new model parameterized by $\tilde\theta$. We provide key 
takeaways for each feedback-update pair, summarizing current work and future directions. To identify relevant work in each category of the taxonomy, we used a snowball sampling methodology~\cite{goodman1961snowball} to gather references that pertain to expert feedback incorporation. For clarity, we provide examples for each category to ground our discussion. While each category is its own research direction, we provide a broad, non-exhaustive discussion of how to update models under expert feedback.
Note many references we provide do not consider a domain expert explicitly but can be applied to the feedback that the domain expert provides. While all feedback may not always be immediately actionable, we hope our paper encourages future work to help practitioners address non-technical experts' needs and concerns more efficiently.

\subsection{Domain to Dataset}
\begin{figure}[h!]
\centering
  \includegraphics[scale=0.25]{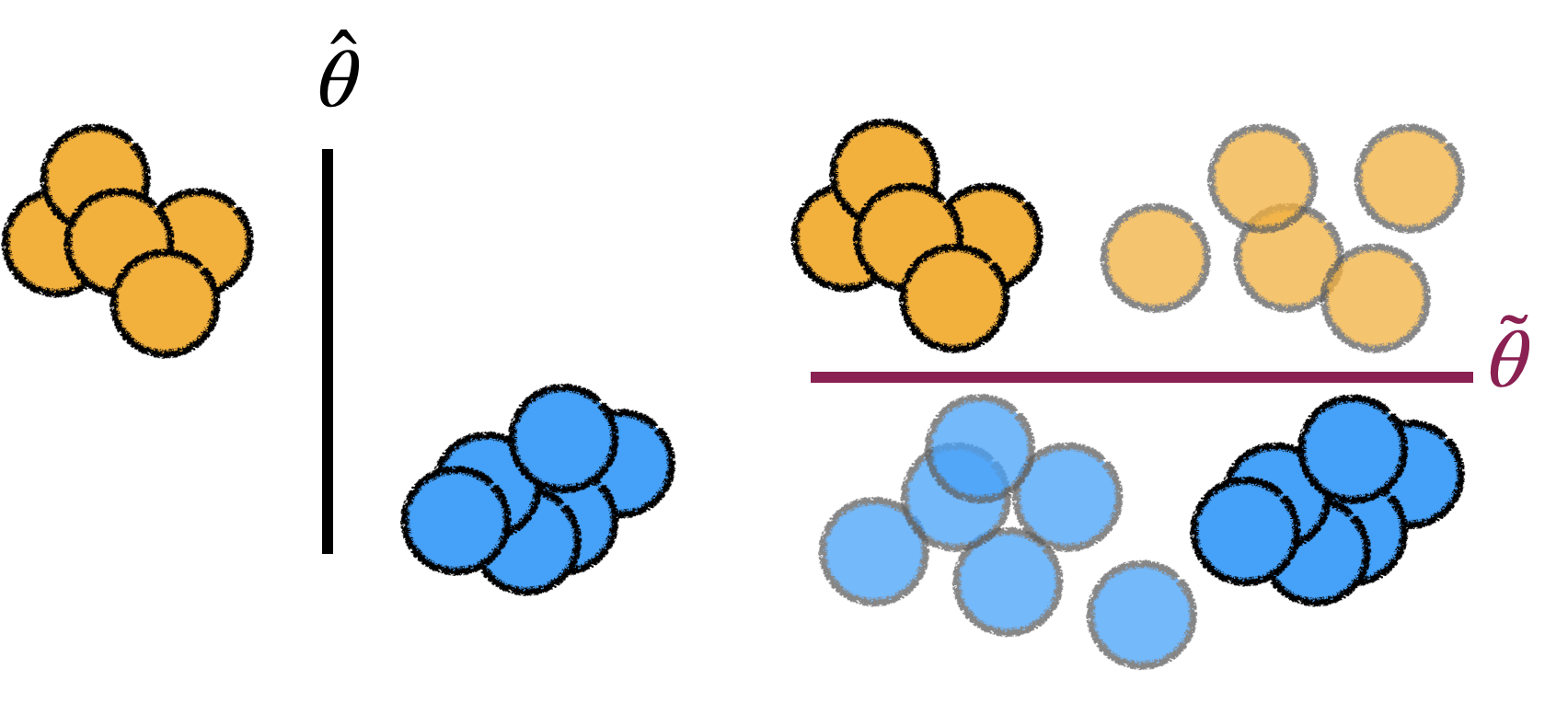}
  \caption{The practitioner generates data, shown in translucent circles, to remove reliance on one feature, after retraining, per the expert's domain-level feedback. The updated dataset, $\mathcal{D}^\prime$, is used to obtain the new parameters $\tilde\theta$~\cite{ratner2017snorkel}.}
  \label{fig:PtoD}
\end{figure}

A dataset can be modified given domain-level feedback. The practitioner can modify the original data or generate new data. If the expert provides domain-level feedback that the model should not rely on a feature, Figure~\ref{fig:PtoD} shows how adding data can lead to a model that satisfies this expert-specified domain-level feedback.

\noindent\textbf{Dataset modifications.} A non-technical expert may specify  feedback which entails a dataset modification and consequently a retraining of the model. One example that naturally suggests a dataset update is dataset preprocessing, so $\mathcal{D}^\prime = h(\mathcal{D})$, where $h$ is a transformation to apply to the original dataset. Suppose the expert suggests that the dataset should be balanced in terms of the sensitive attribute, the practitioner can design an $h$ that achieves this property.
Extensive work has been done on data preprocessing for fairness~\citep{hajian2012methodology,feldman2015certifying, calmon2017optimized}. Another example of domain-level feedback that leads to dataset updates is data augmentation: $\mathcal{D}^\prime = \mathcal{D} \cup \{(h(x_i, y_i))\}_{i=1}^m$ where $h$ is an augmentation scheme. 
After the expert specifies that the model should make the same prediction regardless of image rotation, the practitioner may consider modifying the dataset by adding rotated variants of input images to $\mathcal{D}$~\cite{krell2017rotational}, which allow for more robust inference~\citep{dao2019kernel}.


\noindent\textbf{Data generation.} Other domain-level feedback may not necessarily suggest a dataset modification but rather may prescribe a way to \emph{generate} data to augment the existing dataset, $\mathcal{D}^\prime = \mathcal{D} \cup \{(x_i, y_i)\sim \phi\}_{i=1}^m$, where $\phi$ is a generator specified by the practitioner based on expert feedback. For example, an expert may say that the model should not rely on spurious correlations between people and other objects in images (e.g., using a tennis player to detect the tennis racket). A practitioner may generate counterfactual images from $\mathcal{D}$ that contain and do not contain the people and other objects to decrease reliance on the person in the image~\citep{plumb2021finding}. Generative modelling techniques are useful to create synthetic data that adheres to a property specified by an expert~\citep{howe2017synthetic}. 
For example, to assuage an expert's fairness concerns, a practitioner could generate data under a fairness constraint per \citet{xu2018fairgan}.

\noindent\textbf{Weak supervision.}
Domain-level feedback may be used to handle unlabeled data.
Weak supervision applies to any approach that deals with data where only some is labelled. To leverage additional unlabeled data, 
we can obtain lower-quality labels in an efficient manner~\citep{ratner2019weak}. 
The most common approach is to ask experts to provide higher-level supervision over unlabeled data~\citep{ratner2017snorkel,carmon2019unlabeled,arazo2020pseudo}. For example, an expert may provide an explicit rule stating that: ``All individuals under 18 should have a negative label.'' The practitioner can turn this rule into a pseudo-labelling function $h$, which can be used to leverage large amounts of unlabeled datapoints.

\textit{Takeaway:} Approaches to modifying or generating data are common in the ML literature. However, these methods do not usually involve eliciting supervision from non-technical experts. Future work should focus on prompting experts for domain-level feedback, which could induce dataset changes in settings where collecting more data may be hard: we elaborate in Section~\ref{sec:prompts}.


\subsection{Domain to Loss Function}\label{sec:dtol}
\begin{figure}[h!]
\centering
  \includegraphics[scale=0.25]{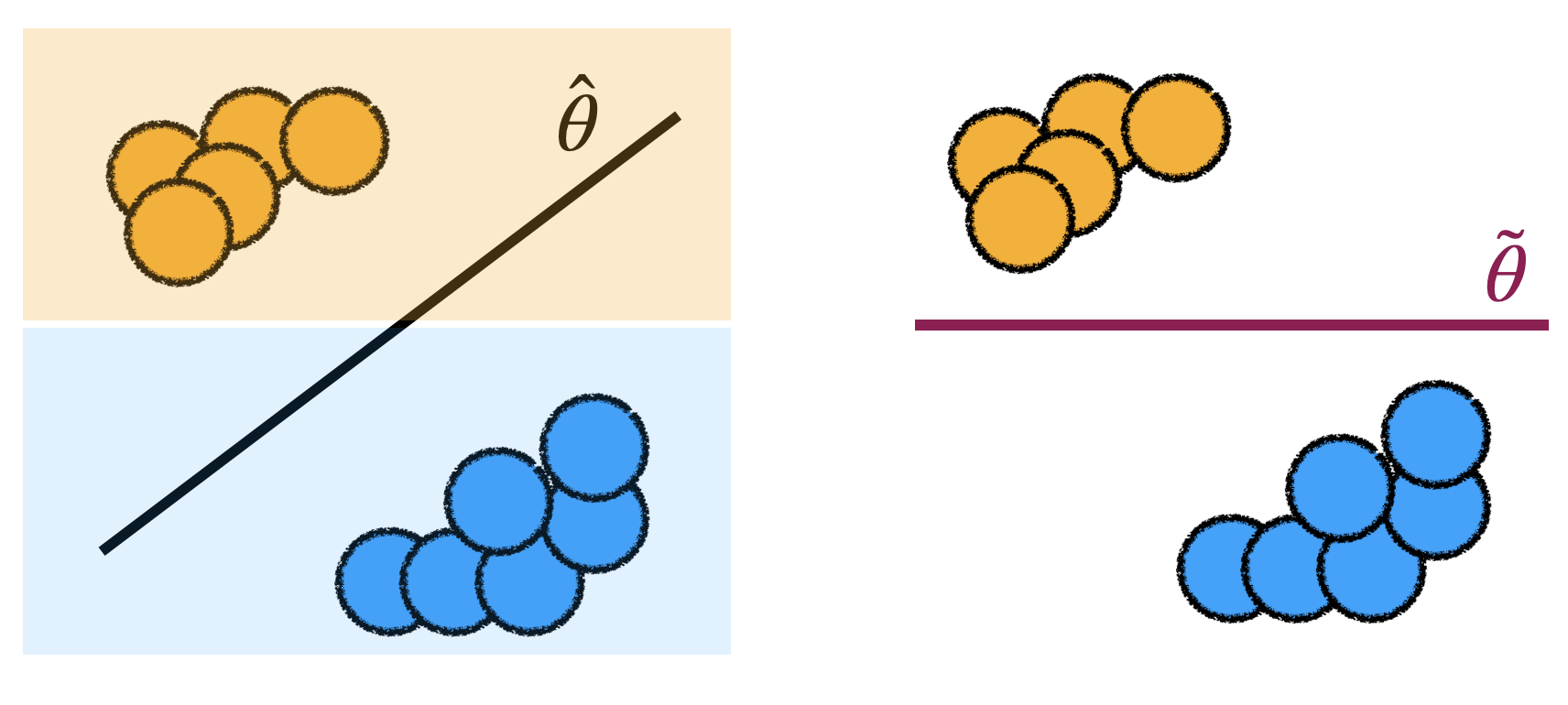}
  \caption{The expert specifies that model behavior should be similar in each color block. The practitioner converts this feedback into a regularizer $\mathcal{R}$, which is used to update $\mathcal{L}^\prime$ and obtain $\tilde\theta$~\cite{slack2020fairness}.}
  \label{fig:PtoL}
\end{figure}

Practitioners can use an expert's domain-level feedback to update the loss function. One example is shown in Figure~\ref{fig:PtoL}.

\noindent\textbf{Constraint specification.} Experts may have properties, beyond performance, that may be well-suited to constrained optimization. There has been a diverse body of work on incorporating fairness~\citep{zafar2017fairness,donini2018empirical,coston2021characterizing} and  interpretability~\citep{lakkaraju2016interpretable, plumb2020regularizing} constraints.
Other concerns experts may express include memory constraints that prohibit models with a large number of parameters~\cite{frankle2018lottery} or run time constraints that require low latency inference~\cite{lin2020mcunet}. For many of these constraints, practitioners may update their loss function to take the following form: $\mathcal{L}^\prime = \mathcal{L} + \lambda\mathcal{R}$, where $\lambda$ is a hyperparameter and $\mathcal{R}$ is the constraint added to the loss. For example, to achieve better fairness outcomes with respect to demographic parity, \citet{slack2020fairness} adds $\mathcal{R} = 1-\frac{1}{|\mathcal{D}_0|}\sum_{x \in \mathcal{D}_0} Pr(f_\theta(x) = 1)$, where $\mathcal{D}_0$ are protected instances. 
Other work improve interpretability by placing constraints on the neighborhood fidelity of the model at each point~\citep{plumb2020regularizing}, which some experts may desire. 

\textit{Takeaway: } The literature demonstrates that it is possible to add property-like constraints to the loss functions. Prior work fails to asks non-technical experts to specify constraints for  two potential reasons. First, providing model information that captures relevant property information may be more difficult to capture than information about specific datapoints. Second, specifying constraints in a usable (e.g., mathematical) format may be challenging for non-technical experts. For example, experts may desire properties that are difficult to convert into a precise statement that can be used to update $\mathcal{L}$ (e.g., societal norms/ethics). 

\subsection{Domain to Parameter Space} \label{sec:parameterudpates2}
\begin{figure}[h!]
\centering
  \includegraphics[scale=0.3]{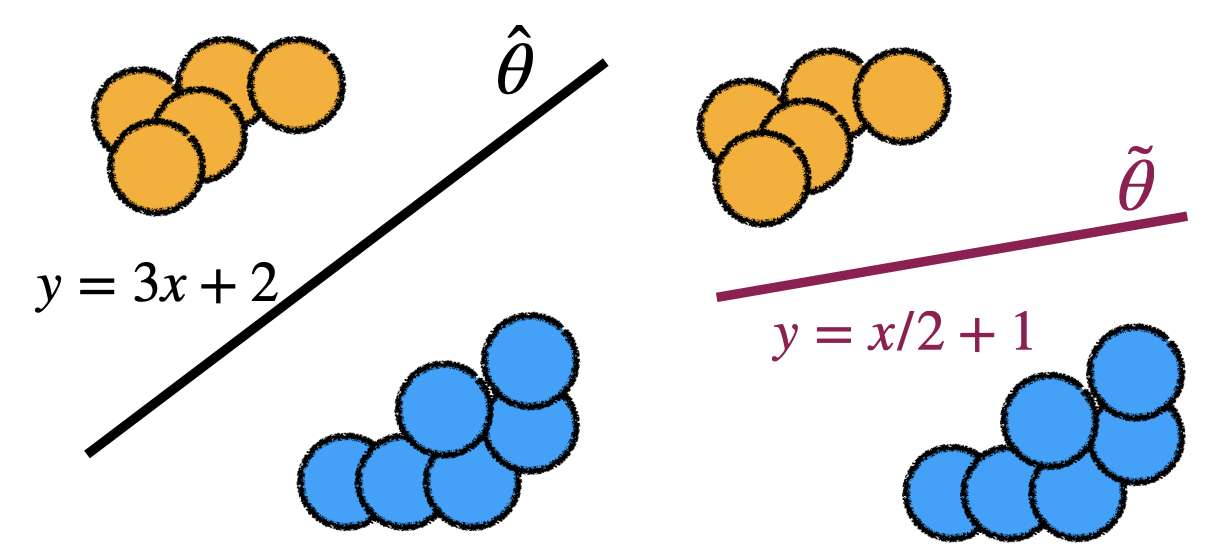}
  \caption{For certain types of models, the expert may be able to directly edit $\hat \theta$, which are the coefficients of the linear model, to obtain their desired model~\cite{wang2021gam}.}
  \label{fig:PtoP}
\end{figure}

Specified domain-level feedback can reduce the parameter space to a set of potential models. In Figure~\ref{fig:PtoP}, we show a simple setting where the expert can directly intervene on model weights.

\noindent\textbf{Model selection.} It may be possible to have the expert provide preferences over \emph{a set} of models. Though there are a number of open questions on how this set is presented (discussed in Section~\ref{sec:prompts}), one solution from \citet{lage2018human} is to show an explanation to an expert, calculate the amount of time it takes for the expert to predict the label, and change the model prior (update) accordingly. The change to the model prior manifests as constraints the practitioner places during fine-tuning: $\Theta^\prime = \{\theta \in \Theta | \; \texttt{local-proxy}(\theta) \leq \epsilon \}$, where \texttt{local-proxy} is a function that calculates the interpretability of $\theta$ and $\epsilon$ is a specified tolerance.
The model prior can also be implicitly changed by imposing complexity constraints on the set of $\Theta$~\citep{rudin2019stop,dziugaite2020enforcing}, which may include constraints on how sparse, interpretable, or smooth the model must be. 

\noindent\textbf{Model editing.}  For some model classes (e.g., simple, transparent models~\citep{wang2021gam,yang2019study}), the expert may \emph{directly} provide feedback that would change the parameters of the model itself without requiring retraining of the model on new data or loss function. For example, \citet{wang2021gam} allow experts to directly change the weights of a generative additive model (GAM) after exposing shape function visualizations and other model properties. In this case, the update to $\Theta$ is trivial, given by $\Theta^\prime = \{ \tilde{\theta} \}$ which is the model that is specified by the expert. These edits on the model might implicitly align the model with unexpressed, desired expert properties, allowing practitioners to avoid the sample-inefficient data collection procedures and the potential difficulties with constraint specification. 

\textit{Takeaway:} While these types of updates traditionally require more technical users, there are increasingly more user-friendly interfaces developed to allow even non-technical experts to edit the model in a more direct manner.

\subsection{Observation to Dataset}
\begin{figure}[h!]
\centering
  \includegraphics[scale=0.24]{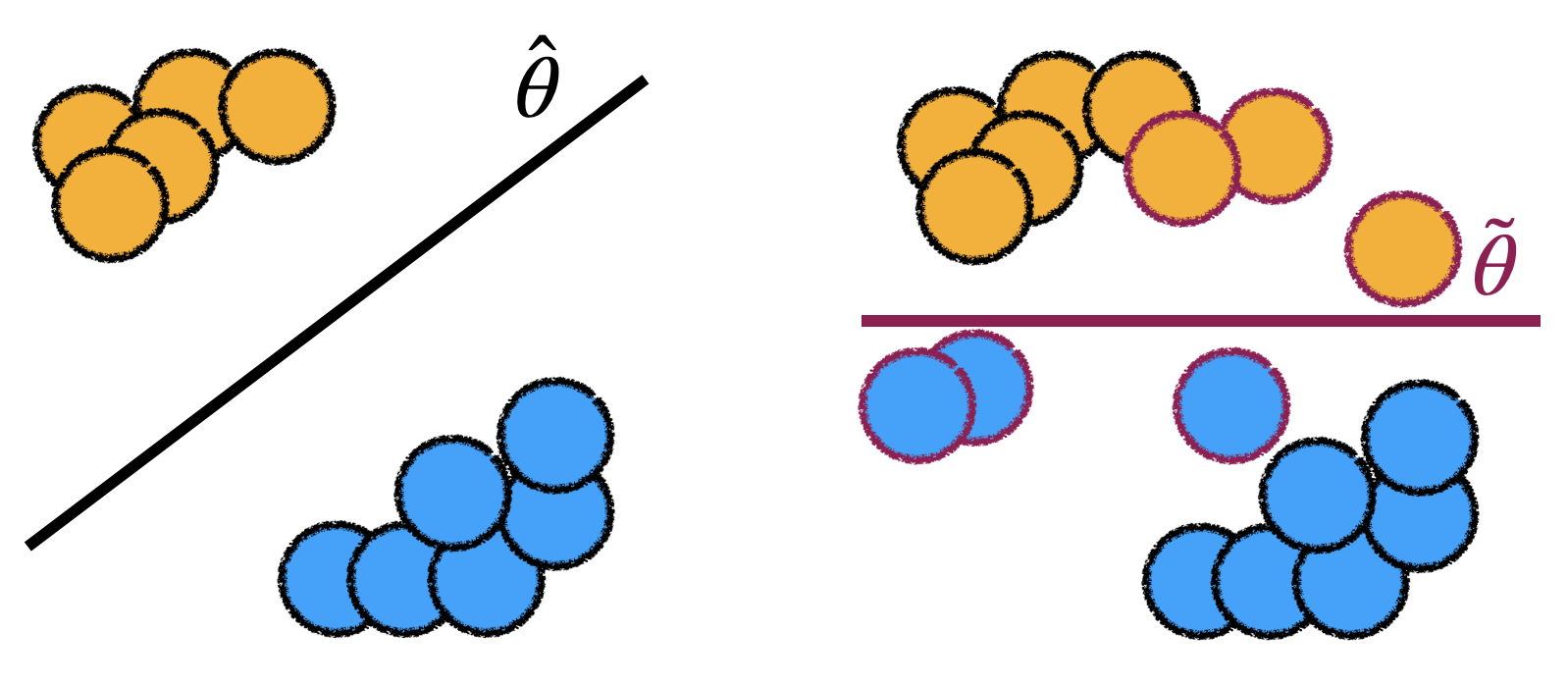}
  \caption{Purple points, specified by the stakeholder, can be added to the dataset to obtain $\tilde \theta$~\cite{irvin2019chexpert}.}
  \label{fig:DtoD}
\end{figure}

The dataset can be directly modified from observation-level feedback via actively collecting data (e.g., asking experts to label selected data points) or passively observing behavior (e.g., collecting data from expert behavior in practice). In Figure~\ref{fig:DtoD}, we show how collecting new data can transform the model into $\tilde\theta$. 

\noindent\textbf{Active data collection.} The field of active learning falls into this category~\citep{settles2009active,gal2017deep}. Traditional work on active learning does not explicitly consider the human in the loop (i.e., the choice of new points to add to $\mathcal{D}$ are selected by a learning algorithm), and it is rather straightforward to use the updated dataset $\mathcal{D}^\prime$ to retrain the model. 
Data collection may transform $\mathcal{D}$ into $\mathcal{D}^\prime$ by asking the expert to approve the weight placed on~\citep{bourtoule2021machine} or to provide a label for a datapoint~\citep{kaushik2019learning}. 
Experts may also review new datapoints, where each new datapoint $x$ is selected by some heuristic (e.g., high uncertainty regions) and the corresponding $y$ is specified by the expert~\citep{wan2020human,fanton2021human}. 
Some work considers collecting multiple labels from various experts for each $x$~\cite{peterson2019human}.
Recently, active learning has been studied alongside model transparency, specifically using explanations to assist experts with choosing which points to add to $\mathcal{D}$~\citep{ghai2020explainable}. \citet{cabrera2021discovering} propose an extensive visual analytics system that allows experts to verify and produce examples of crowd-sourced errors, which can be thought of as additional data.



\noindent\textbf{Passive observation.} Instead of asking the expert to provide labels on additional pieces data, the practitioner could also collect data via expert demonstrations. 
Inverse decision theory argues for observing human decisions to learn their preferences~\citep{swartz2006inverse}. One can simply observe a non-technical expert's behavior to generate more data to include in $\mathcal{D}$~\cite{hertwig2009description}. 
For example, a data scientist might choose to wait for radiologists to see more patients before updating the dataset and  model~\citep{irvin2019chexpert}.
While this may not be the most efficient way to perform model updates, \citet{laidlaw2021uncertain} find that forcing humans to make decisions under uncertainty can lead to better preference learning. Note that the online learning community has extensively studied how to incorporate expert knowledge into learning statistical models based on a sequence of observation~\citep{littlestone1994weighted,hannan1957approximation,hoi2021online}.

\textit{Takeaway: } Active data collection may not immediately appear to be a reasonable feedback to expect from non-technical experts. Compared to the traditional crowdworkers used in the ML literature, non-technical experts like lawyers and regulators may not have the time to extensively label data that is required for some powerful deep learning methods. However, we do not exclude this work because it may still be desirable to collect data from non-technical experts (e.g., in settings with limited amounts of data) or collect observation-level feedback in a passive manner.  

\subsection{Observation to Loss Function}

\begin{figure}[h!]
\centering
  \includegraphics[scale=0.25]{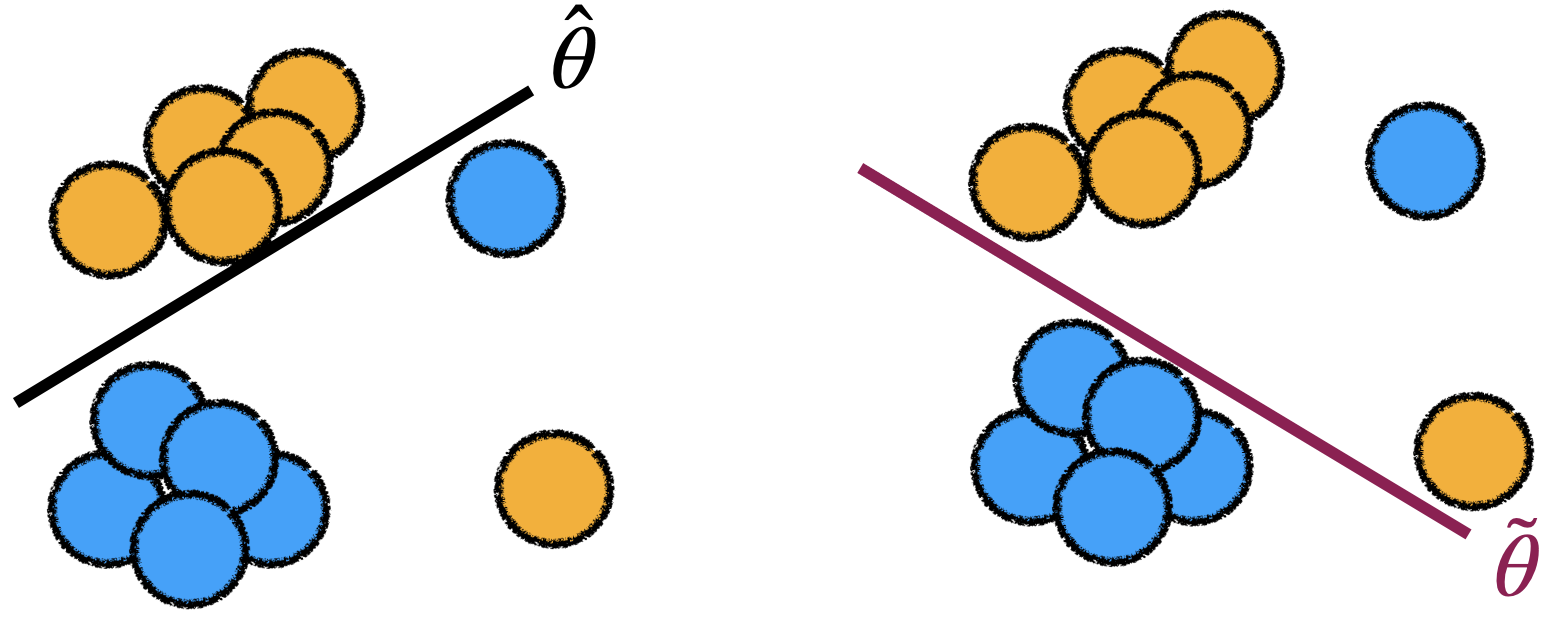}
  \caption{The expert specifies that cost of mislabeling a yellow point is higher than the cost of mislabeling a blue point. Retraining with $\hat{\mathcal{L}} = \mathcal{L} + \mathcal{R}(s_i)$, where $s_i$ is the cost of error for each point, yields $\tilde \theta$~\cite{elkan2001foundations}.}
  \label{fig:DtoL}
\end{figure}

We identify two ways to collect additional information to learn and integrate a new function into the loss. In Figure~\ref{fig:DtoL}, we show that practitioners can collect contextual information in the form of error costs to edit the loss function and retrain the model. 

\noindent\textbf{Collecting contextual information.} 
Experts may have contextual information they wish to share with practitioners.
While there is not a unified way for practitioners to incorporate contextual information, it is often used for a loss function update. 
The contextual information might be provided for some subset of points if collecting it for all datapoints is expensive~\cite{elkan2001foundations,miao2021incorporating}. 
In general, contextual information can be used to constrain the model behavior~\citep{dekel2017online,zhao2017learning}.

\citet{tseng2020fourier,weinberger2020learning} both regularize feature attributions to ensure that the explanations from models better align with expert expectations. They let $S$ be the feature-attribution explanations for each training point, as specified by the expert, and add the following to the loss:
$\mathcal{R}(S) \coloneqq \sum_{x_i \in \mathcal{D}} \sum_{j \in d} g(\theta;x_i)_j - s_{ij}$ where $S \in \mathcal{R}^{n\times d}$ is the desired attribution and $g$ is the explanation function using our model $\theta$. 
\citet{koh2020concept} proposed concept bottleneck models (CBMs) as a way to incorporate pre-defined concepts into a supervised learning procedure: concepts are semantically meaningful pieces of information used in a discriminative model to perform prediction~\citep{kim2018interpretability,ghorbani2019towards}. 
Their supervised approach maps raw inputs ($x$) to concepts ($c$), and then maps concepts ($c$) to targets ($y$). An intermediate layer of a neural network can also be selected as the CBM, where the layer's activations should be aligned with concepts when training~\citep{ross2017right,lage2020learning}.

Furthermore, expert-specified contextual information can be used in other clever ways. \citet{vapnik2015learning} use \textit{privileged information} for each input to accelerate learning of a support vector machine.
\citet{abe2004iterative} use \textit{misclassification costs} to find a weighted loss function that improves model performance under class imbalance.
\citet{adel2018discovering} use \textit{generative factors} to improve deep representation learning. 
\citet{hind2019ted} use \textit{explanations} to partition classes into subclasses for more accurate models downstream.

\noindent\textbf{Constraint elicitation.} The practitioner may choose to parameterize a constraint using observation-level feedback. Analogous to metric learning~\cite{davis2007information}, this can be done by learning the hyperparameters of a constraint from expert feedback about individual points~\citep{yaghini2021human} or by building a function from expert observations~\citep{jung2019algorithmic,wang2019empirical,ilvento2020metric}. The learned metric, $\mathcal{R}$, is then appended onto the existing loss function. For example, some have attempted to learn an individual fairness constraint after receiving pairwise judgements from experts, who specify if two individuals should be treated the same or not~\citep{jung2019algorithmic,wang2019empirical}. 
Practitioners can also constrain intermediate model representations~\citep{hilgard2021learning} or edit a model's representation~\citep{santurkar2021editing}, 
which entails experts selecting exemplar training points that should have a similar representations to a test point.

\textit{Takeaway: } The aforementioned approach of constraint specification (see Section~\ref{sec:dtol}) bears similarity to both approaches we discuss here. The main difference between using constraint specification and collecting contextual information is that contextual information is specified for each datapoint, which may be easier for an expert to provide generally. Although constraint elicitation techniques circumvent the potential difficulty of constraint specification, these techniques require more creativity on how the individual pieces of observation-level feedback can be combined. 
We note that there should be further work on organizing these types of approaches, as they likely will be domain specific. 


\subsection{Observation to Parameter Space} 
\begin{figure}[h!]
\centering
  \includegraphics[scale=0.3]{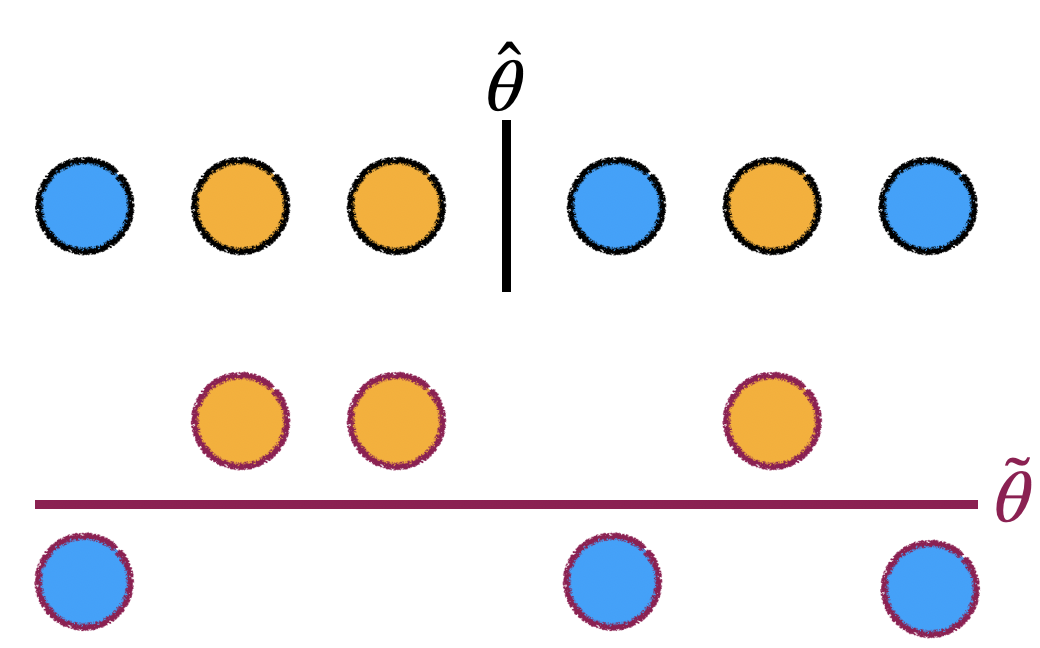}
  \caption{Having the expert add a new feature to the dataset allows the datapoints to become linearly separable, which leads to a more accurate model~\cite{bakker2021beyond}.}
  \label{fig:DtoP}
\end{figure}

In some cases, the parameter space may not be rich enough to find a suitable model that fits the dataset well~\citep{laidlaw2021uncertain}. Additional data containing modified features can be used to alter parameter space. In Figure~\ref{fig:DtoP}, we show how adding a feature to a dataset allows $\tilde{\theta}$ to better separate the blue and yellow points.

\noindent\textbf{Feature modification.}
This type of update manifests as a change to the parametric form of the model, which now makes predictions on the changed datapoints. \citet{correia2019human, roe2020feature} use experts to select a \textit{subset} of features to use for prediction (e.g., to bias the model away from using spurious correlations). \citet{bakker2021beyond} sequentially add features to a dataset to achieve fairness goals.
In these works, the original parameter space may be all 2D models, $\Theta = \mathbb{R}^2$, but the updated parameter space after feature acquisition would be $\Theta^\prime = \mathbb{R}^3$. Moreover, there is a plethora of work on feature selection that implicates the selected model class~\cite{greenland1989modeling,george2000variable}. There have been works where experts can suggest the model class based on their interpretability needs: the less complex the parameter space, the more interpretable the model~\cite{rudin2019stop,dziugaite2020enforcing}.

\textit{Takeaway:} Working with experts to modify features might be particularly helpful earlier in  model development. We note that this type of update is already commonly done in practice.

\section{Illustrative Example}

To contextualize our feedback-update taxonomy, consider a lawyer, a non-technical domain expert who provides feedback on a model that predicts likelihood of recidivism~\citep{dressel2018accuracy}. We provide three scenarios to illustrate different potential interactions between the practitioner and lawyer.

\smallskip
\noindent \textit{Scenario \#1: Lawyer says ``The dataset used does not seem representative of the population. Specifically, there are not enough individuals from a minority group.'' } Since the lawyer recommends expanding the dataset, this feedback-update naturally falls in the Observation to Dataset category under Active data collection. The practitioner can try a few techniques from this category: The practitioner could collect more data from the minority group, but that might be infeasible and expensive. The practitioner thus may decide to reweight the dataset so that the dataset is balanced across subgroups~\cite{zhao2019metric}. This is a straightforward case where modifying the dataset is a good way to handle the expert feedback. 

\smallskip
\noindent \textit{Scenario \#2: Lawyer says ``This factor usually has limited importance for predicting recidivism.''} The practitioner identifies that this is a domain-level feedback, narrowing down the set of approaches to making an update. One way the practitioner can take this feedback into account is to generate data so that the suggested feature is not correlated with outcome (Domain to Dataset)~\cite{xu2018fairgan}. Another way the practitioner can incorporate this feedback is to regularize for gradients of the unimportant factor (Domain to Loss function)~\cite{dimanov2020you}. 

\smallskip
\noindent \textit{Scenario \#3: Lawyer says ``I have a general sense of what the model should not do.''} In this case, it is not clear whether the expert would like to provide domain- or observation-level feedback. The practitioner should work with the expert to determine which approach is best, since the expert does not have an actionable piece of feedback. If the lawyer is willing to answer a set of carefully designed questions, one approach the practitioner can try is constraint elicitation by asking a series of pairwise questions to learn the lawyer's metric (Observation to Loss function)~\citep{jung2019algorithmic}. If the lawyer has less time and the model is a simple model class, the practitioner could have the lawyer try model editing (Domain to Parameter space)~\citep{wang2021gam}.
\smallskip

In these various situations, the taxonomy summarizes the ways that the lawyer's feedback could be translated and incorporated into the model, providing a way for the practitioner to begin reasoning about the choices they have at hand along the two axes. However, the feedback that the expert provides may not be immediately actionable. As a result, this interaction will require iterative communication between practitioner and expert. Such interaction can be aided by choices of model information, prompts to the expert, and consideration of how to best update the model, which we discuss next as open questions.

\section{Open Questions}
\label{sec:open}
Through our review of practitioner-expert interactions, we find there are multiple important and exciting open questions. While our taxonomy highlights many ways to convert feedback to updates, there are still open questions on how practitioners should prompt and collect feedback from non-technical experts (Section~\ref{sec:prompts}) and how practitioners can decide the best type of update to perform given feedback (Section~\ref{sec:troubleupdate}). 
We discuss the potential \emph{algorithmic} and \emph{participatory} innovations needed to improve our feedback-update pipeline: collaborative effort between multiple communities is imperative to increase non-technical expert involvement in model development.



\subsection{Difficulties with Feedback Collection}
\label{sec:prompts}
There are many complexities that arise when practitioners collect feedback from non-technical experts. We motivate two: \textit{prompting for feedback} and \textit{engaging with multiple experts}.

\subsubsection{Prompting for Feedback} \label{sec:prompting}

\smallskip
\begin{center}
    \textit{What \textbf{information} about the model should the practitioners present to the expert? What type of \textbf{prompt} should a practitioner choose?}
\end{center}
\smallskip

In the process of prompting experts for feedback, practitioners need to decide how and what information about the model needs to be shown to experts. There has been initial work to develop interfaces to visualize trade offs between multiple objectives~\citep{yu2020keeping}, to allow experts to explore model behavior~\cite{cai2019human,brown2019toward,katell2020toward}, or to interact with white-box proxies~\cite{yang2019study}. There has been a flurry of ML monitoring work, which provides dashboards to assess models in production~\citep{karumuri2021towards,rabanser2019failing}. While these techniques try to bridge the communication gap between practitioners and experts, these tools seldom provide adequate remediation for experts to express their preferences, and for practitioners to incorporate those preferences.


\begin{figure*}[htb]
\centering
    \begin{subfigure}[b]{0.24\linewidth}
        \includegraphics[width=0.95\textwidth]{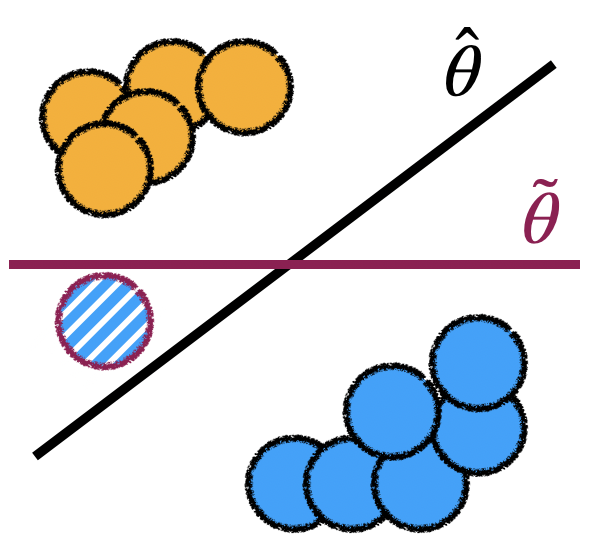}
        \caption{\scriptsize Open-Ended Data}
        \label{open-data}
    \end{subfigure}
    \begin{subfigure}[b]{0.24\linewidth}
        \includegraphics[width=0.95\textwidth]{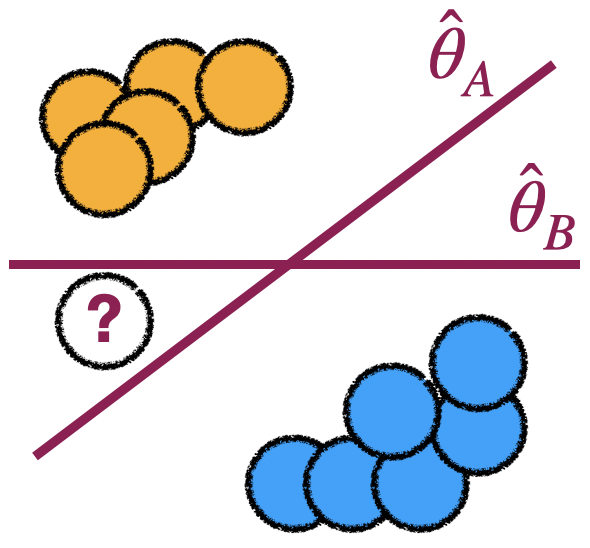}
        \caption{\scriptsize Forced-Choice Data}
        \label{choice-data}
    \end{subfigure}
    \begin{subfigure}[b]{0.24\linewidth}    
        \includegraphics[width=0.95\textwidth]{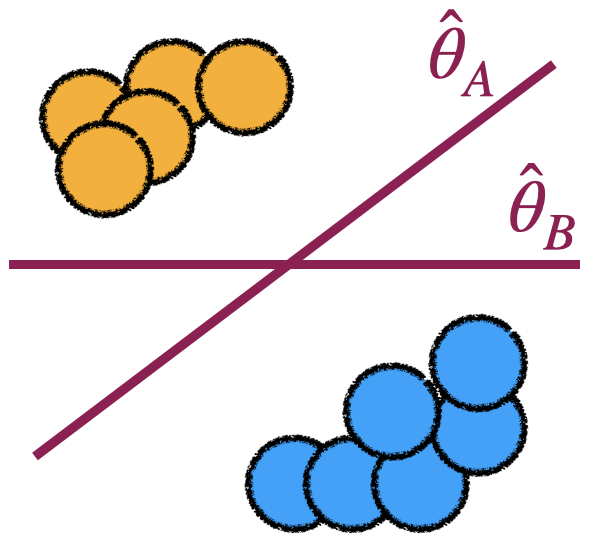}
        \caption{\scriptsize Forced-Choice Property}
        \label{choice-prop}
    \end{subfigure}
    \begin{subfigure}[b]{0.24\linewidth}
        \includegraphics[width=0.95\textwidth]{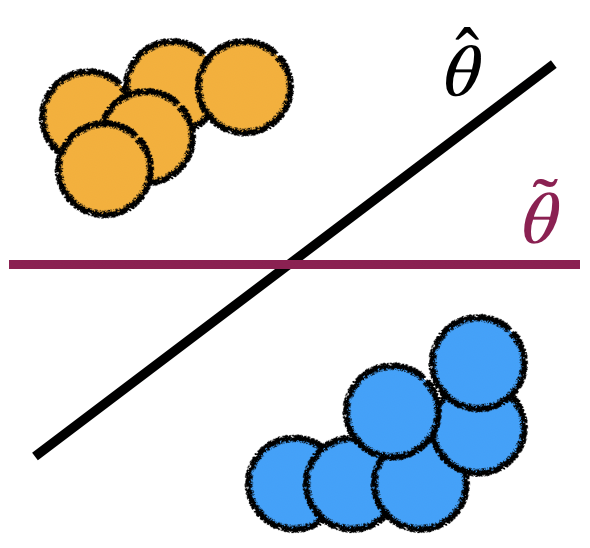}
        \caption{\scriptsize Open-Ended Property}
        \label{open-prop}
    \end{subfigure}%
    \caption{We illustrate four different combinations of feedback type and prompt styles that a practitioner might ask from an expert. (a) The practitioner presents a model parameterized by $\hat\theta$; the expert specifies the addition of the striped point with a purple border to the blue class. The induced change by this feedback yields a new model parameterized by $\tilde\theta$. (b) The practitioner presents the expert with a forced-choice to provide a label for the point denoted with a question mark. Depending on the selected label, the induced classifier may be parameterized by $\hat\theta_A$ or $\hat\theta_B$. (c) The practitioner presents the expert with a forced-choice of two properties: enforce max-margin between two classes ($\hat\theta_A$) or avoid reliance on the vertical feature ($\hat\theta_B$). (d) The expert specifies a property that the model should not use the vertical feature to make predictions, inducing the same classifier as the model parameterized by $\hat\theta_B$ in (b) and (c).}\label{fig:feedback}
\end{figure*}

Given experts might want to provide different types of feedback~\cite{honeycutt2020soliciting}, an important open question is how practitioners should guide non-technical experts via \emph{prompts} to provide specific feedback. Two such prompt styles include \emph{open-ended feedback} and \emph{forced-choice feedback}. In Figure~\ref{fig:feedback}, we differentiate how these feedback types can be presented with a simple example. However, challenges arise both in dealing with the types of prompts~\cite{holzinger2016interactive} and deciding how to word the prompt.

In open-ended feedback, experts are unconstrained in the information they can provide. 
Thus, the onus is on experts to identify relevant aspects of model behavior on which to opine and to do so in a usable manner~\cite{robertson2020if}. 
Open-ended feedback is more challenging to use, insofar as the feedback may have little to no implication on the model. 
For example, subjective or qualitative feedback (e.g., measures of confidence or trust~\cite{wang2019designing,jacovi2021formalizing}) may be difficult to translate into downstream model updates. 

Forced-choice feedback is feedback presented to the expert in the form of options, where the expert is forced to pick either one, multiple, or rank the options. 
There is a related question of how to select, visualize, and present options. For example, in the growing body of work that studies a $\epsilon$-Rashomon set of good models~\cite{breiman2001statistical},
there are currently only heuristics to identify this set of models,\footnote{This may be done by setting multiple random seeds~\cite{semenova2019study}, optimizing under constraints~\cite{bhatt2020counterfactual}, or subsampling from the dataset~\cite{marx2020predictive}.} and it remains unclear how to visualize this set. 

\noindent\textbf{Algorithmic Innovation.} For open-ended feedback, new methods can let experts better traverse model behavior to improve holistic understanding of the model. For forced-choice feedback, future work can build methods to understand the order in which the queries should be asked and what set of options to provide. 

\noindent\textbf{Participatory Innovation.} The practitioner's choice of content and presentation of model information will affect the expert's downstream feedback~\citep{schnabel2018improving}. Additional user studies are needed to understand the most appropriate pairings of model transparency, prompts, and interfaces to guide the development of useful tools to involve non-technical experts, recognizing potential challenges \citep{weller2019transparency,coyle2020explaining,zerilli2022transparency}.

\subsubsection{Engaging with Multiple Experts}
\label{sec:agg_feed}
\smallskip
\begin{center}
    \textit{How does a practitioner perform an update after receiving
    feedback from \textbf{multiple} experts?}
\end{center}
\smallskip

When non-technical domain experts are considered in our pipeline, there may be multiple, diverse experts who want to express feedback on a model.
Thus, it is important to expand our taxonomy to handle multiple domain experts.
In an ideal world, there would be an easy way to incorporate feedback from multiple domain experts into models: this is related to participatory ML~\citep{paml}, a burgeoning community democratizing the ability for all to interact with models. 

Practitioners 
should combine feedback from multiple domain experts carefully, perhaps using an aggregation mechanism~\cite{kahng2019statistical,lee2019webuildai}. 
The combinatorial auctions literature has long studied how to best elicit preferences when the number of options is combinatorial~\citep{cramton2004combinatorial,cramton2007overview}. 
Every feedback aggregation mechanism is value-laden, as some mechanisms might prefer one expert over others and some tie-breaking mechanisms might under-represent some experts~\cite{freeman2015general,ward2021value,levin1995introduction,azari2014statistical}.
Even if practitioners had a sensible feedback aggregation mechanism, settling incongruities in feedback is difficult, as there might be conflicting pieces of feedback from multiple, diverse domain experts~\cite{brandt2012computational,halfaker2020ores,kahng2019statistical,lee2019webuildai,robertson2020if}.

\noindent\textbf{Algorithmic Innovation.} Developing methods to aggregate feedback from multiple domain experts will require a framework that not only considers the heterogeneous feedback types but also permits various aggregation schemes, which combine potentially contradictory feedback into a collective piece of feedback for the practitioner to incorporate~\citep{gordon2022jury,de2021leveraging}.


\noindent\textbf{Participatory Innovation.} 
New methods should consider ways to efficiently elicit feedback at scale from diverse domain experts and mechanisms to make explicit the value-laden aggregation done when ironing out potential contradictions in collected feedback.

\subsection{Navigating Selection of Model Updates}\label{sec:troubleupdate}
\label{sec:unify_update}
\smallskip
\begin{center}
    \textit{How can a practitioner best collect feedback to make the most \textbf{impactful} update to the model?}
\end{center}
\smallskip

Non-technical domain expert feedback is difficult to collect~\cite{lage2018human,hilgard2021learning}. As a result, practitioners need to ensure that, given a single piece of feedback, the update they make has a large impact on the model. To measure the impact of an update, there are a few open questions to address. First, it is not clear what \emph{impactful} means. 
One naive way is to measure the amount of feedback needed to affect a specified change in the model. 
However, this ignores how different types of feedback may have different costs of collection and may be amenable to different types of updates. 
A practitioner may need to choose between feedback/update pairs because multiple update types could be interchangeable (i.e., a practitioner can get the \emph{same} parameters $\tilde \theta$) given a piece of feedback. 
For example, a domain expert's fairness goals can be achieved by clever sampling from the dataset~\cite{dong2018imbalanced} or by adding constraints to the loss~\citep{zafar2017fairness}. Others have connected loss function and dataset changes using optimization~\cite{suggala2018connecting}, Bayesian methods~\cite{khan2021knowledge}, and group theory~\cite{chen2020group}.

\noindent\textbf{Algorithmic Innovation.} 
There is much work to be done in comparing update types to understand what updates are easier to make for what feedback. Understanding the conditions under which practitioners can use either interchangeably will be important. From a technical perspective, practitioners can estimate the complexity of an update (e.g., as in \citet{laidlaw2021uncertain}).

\noindent\textbf{Participatory Innovation.} To determine which type of update should be performed, an important factor to consider is the effort required to collect each kind of feedback. Domain-level feedback may require focus groups and workshops~\cite{cheng2021soliciting}. Observation-level feedback might be faster to collect (e.g., large scale data-labeling platforms~\cite{ratner2017snorkel}), making it easy to collect 
dataset changes or additions~\citep{dean2020recommendations}. 
Creating efficient ways for practitioners to collect domain-level feedback may reduce the complexity of a loss function update, which might be preferable to dataset updates in resource-constrained settings.

\section{Conclusion}
As machine learning is increasingly deployed in key societal settings, there is a growing need to incorporate domain expert preferences into models.
Practitioners need mechanisms to gather and incorporate feedback from non-technical experts into the models they develop. 
In this work, we studied the interaction between practitioner and expert on how feedback can be collected and then used to update the model itself.
We propose a taxonomy to convert feedback from an expert, who can provide domain- or observation-level feedback, into model updates, which changes the dataset, loss function, or parameter space. 
We conclude with concrete open questions that pertain to prompting for feedback, engaging with diverse feedback, and selecting the update type appropriately. 
We implore the community to study how to best incorporate domain expertise into the machine learning development cycle. 



\section*{Acknowledgments}
The authors would like to thank the following individuals for their advice, contributions, and/or support: Kasun Amarasinghe, Varun Babbar, Alex Cabrera, Katherine Collins, Ken Holstein, Nari Johnson, Emma Kallina, Joon Sik Kim, Jeffrey Li, Vera Liao, Leqi Liu, Bradley Love, Gregory Plumb, Charvi Rastogi, and John Zerilli.

UB acknowledges support from DeepMind and the Leverhulme Trust via the Leverhulme Centre for the Future of Intelligence (CFI), and from the Mozilla Foundation. AW acknowledges support from a Turing AI Fellowship under grant EP/V025279/1, The Alan Turing Institute, and the Leverhulme Trust via CFI. AT acknowledges support from the National Science Foundation grants IIS1705121, IIS1838017, IIS2046613, IIS2112471, an Amazon Web Services Award, a Facebook Faculty Research Award, funding from Booz Allen Hamilton Inc., and a Block Center Grant. Any opinions, findings and conclusions or recommendations expressed in this material are those of the author(s) and do not necessarily reflect the views of any of these funding agencies.

\scriptsize
\bibliographystyle{abbrvnat}
\bibliography{acmart}

\end{document}